\pdfoutput=1
\documentclass[letterpaper, 10 pt, conference]{ieeeconf}  

\IEEEoverridecommandlockouts                              

\title{\LARGE \bf
TopoNav: Topological Navigation for Efficient Exploration in Sparse Reward Environments
}

\usepackage{cite}
\usepackage{amsmath,amssymb,amsfonts}
\DeclareMathOperator*{\argmax}{arg\,max}
\usepackage[ruled,vlined,linesnumbered]{algorithm2e}
\usepackage{algpseudocode} 
\usepackage{graphicx}
\usepackage{textcomp}
\usepackage{subfig}
\usepackage{subcaption}
\usepackage{multirow}
\usepackage{tikz}
 \def\BibTeX{{\rm B\kern-.05em{\sc i\kern-.025em b}\kern-.08em
    T\kern-.1667em\lower.7ex\hbox{E}\kern-.125emX}}

\usepackage{colortbl}
\usepackage{soul}
\usepackage{float}
\usepackage[colorlinks=true, allcolors=blue]{hyperref}%

\let\labelindent\relax
\usepackage{enumitem}
\newlist{inlineroman}{enumerate*}{1}
\setlist[inlineroman]{itemjoin*={{, and }},afterlabel=~,label=\roman*.}

\newlist{Inlineroman}{enumerate*}{1}
\setlist[Inlineroman]{itemjoin*={{, and }},afterlabel=~,label=\Roman*.}

\setlist[enumerate]{nosep}
\setlist[itemize]{nosep}
\definecolor{mintgreen}{rgb}{0.6, 1.0, 0.6}
\definecolor{pastelviolet}{rgb}{0.8, 0.6, 0.79}
\definecolor{peridot}{rgb}{0.9, 0.89, 0.0}
\definecolor{richbrilliantlavender}{rgb}{0.95, 0.65, 1.0}
\definecolor{robineggblue}{rgb}{0.0, 0.8, 0.8}

\definecolor{green}{rgb}{0.1,0.1,0.1}

\author{Jumman Hossain$^{1}$, Abu-Zaher Faridee$^{1,2}$, Nirmalya Roy$^{1}$, Jade Freeman$^{3}$, Timothy Gregory$^{3}$, and Theron Trout$^{4}$%
\thanks{$^{1}$Authors are with the Dept. of Information Systems, University of Maryland, Baltimore County, USA. {\tt\scriptsize \{jumman.hossain, faridee1, nroy\}@umbc.edu}}%
\thanks{$^{2}$Author is with Amazon Inc. USA. {\tt\scriptsize abufari@amazon.com}}%
\thanks{$^{3}$Authors are with DEVCOM Army Research Lab, USA. \newline \hspace*{1em}{\tt\scriptsize \{jade.l.freeman2.civ, timothy.c.gregory6.civ\}@army.mil}}%
\thanks{$^{4}$Author is with Stormfish Scientific Corporation. \newline \hspace*{1em}{\tt\scriptsize 
theron.trout@stormfish-sci.com}}%
}

\begin{document}

\maketitle
\thispagestyle{empty}
\pagestyle{empty}




\begin{abstract}


Autonomous robots exploring unknown environments face a significant challenge: navigating effectively without prior maps and with limited external feedback. This challenge intensifies in sparse reward environments, where traditional exploration techniques often fail. In this paper, we present \textit{TopoNav}, a novel topological navigation framework that integrates active mapping, hierarchical reinforcement learning, and intrinsic motivation to enable efficient goal-oriented exploration and navigation in sparse-reward settings. \textit{TopoNav} dynamically constructs a topological map of the environment, capturing key locations and pathways. A two-level hierarchical policy architecture, comprising a high-level graph traversal policy and low-level motion control policies, enables effective navigation and obstacle avoidance while maintaining focus on the overall goal. Additionally, \textit{TopoNav} incorporates intrinsic motivation to guide exploration towards relevant regions and frontier nodes in the topological map, addressing the challenges of sparse extrinsic rewards. We evaluate \textit{TopoNav}  both in the simulated and real-world off-road environments using a Clearpath Jackal robot, across three challenging navigation scenarios: goal-reaching, feature-based navigation, and navigation in complex terrains. We observe an increase in exploration coverage by 7-20\%, in success rates by 9-19\%, and reductions in navigation times by 15-36\% across various scenarios, compared to state-of-the-art methods.

\end{abstract}

\section{Introduction}

\begin{figure}[!htb]
    \centering
    \includegraphics[width=0.45\textwidth]{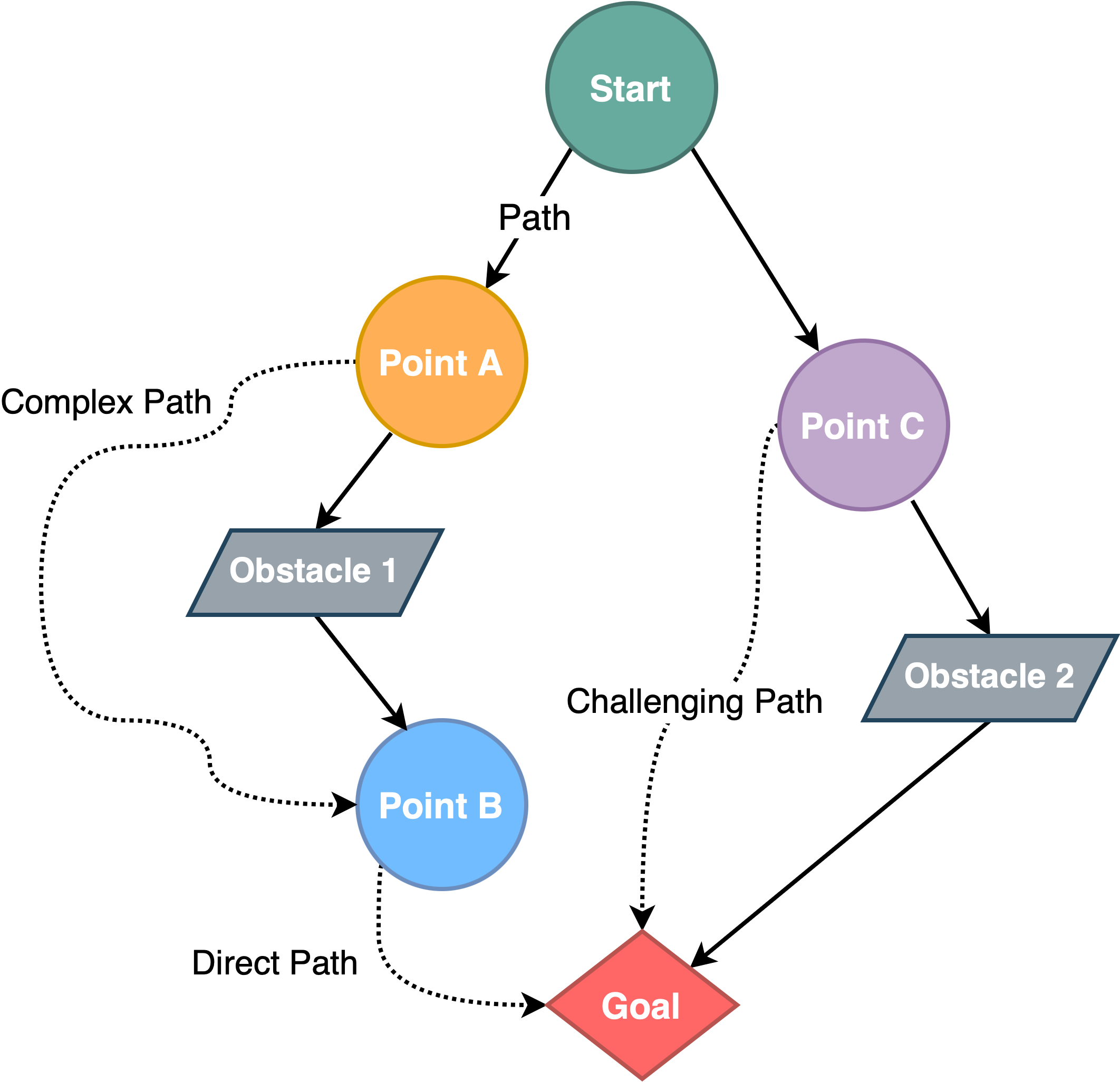}
    \caption{\small 
    \textit{TopoNav Navigation Strategies}: 
The navigation begins at the Start node (green circle) and progresses through designated subgoals—Point A (initial decision point), Point B (complex navigation subgoal), and Point C (alternative challenging subgoal)—toward the Goal (red diamond). The routes illustrate \textit{TopoNav}'s strategy: solid lines represent direct paths to subgoals, a dashed line marks a complex detour around Obstacle1, and a dotted line indicates a potential route for challenging maneuvering near Obstacle2. This diagram shows the robot's strategic navigation from start to finish, highlighting its decision-making and adaptability in outdoor environments with diverse navigational challenges. A real-world scenario is presented in Fig. \ref{fig:obstacle_experiment}}
    

    \label{fig}
\end{figure}
Autonomous robot navigation in unknown, unstructured environments poses significant challenges, particularly in the absence of prior maps and reliable localization ~\cite{cadena2016past,durrant2006simultaneous}. In such scenarios, robots must efficiently explore the environment, build accurate representations, and make intelligent decisions to reach their goals, often with limited computational resources and sparse feedback ~\cite{chen2019learning,niroui2019deep}. Traditional approaches, such as simultaneous localization and mapping (SLAM) ~\cite{grisetti2010tutorial} and sampling-based planning ~\cite{elbanhawi2014sampling}, rely heavily on geometric representations and struggle to adapt to the uncertainties and dynamicity of real-world environments. Recent advancements in deep reinforcement learning (RL) have shown promise in enabling robots to learn complex navigation policies directly from raw sensory inputs ~\cite{tai2017virtual,kahn2018self}. However, most RL-based approaches suffer from sample inefficiency, poor generalization to unseen environments, and difficulties in handling sparse reward signals ~\cite{eysenbach2019search}. Hierarchical RL methods ~\cite{levy2019learning,zhang2020generating} attempt to address these issues by learning multi-level policies, but often rely on handcrafted state spaces, pre-defined sub-goals, and task-specific reward shaping ~\cite{nachum2018data}, limiting their autonomy and adaptability. Topological mapping ~\cite{kuipers2000spatial,boal2014topological} is a promising approach for efficient navigation in large-scale environments. By representing the environment as a graph of discrete places and their connectivity, topological maps provide a compact and flexible representation that scales well with the size of the environment ~\cite{blochliger2018topomap}. However, most topological mapping approaches rely on predefined place recognition methods ~\cite{lowry2016visual} or assume a fixed set of landmarks ~\cite{savinov2018semi}, making them sensitive to handling ambiguous visual appearances and changes in the environment. They often struggle to adapt to dynamic and unstructured environments, where the appearance of landmarks may change significantly over time.

In this paper, we introduce \textit{TopoNav}, a novel topological navigation framework that integrates active mapping, hierarchical reinforcement learning, and intrinsic motivation to enable efficient and autonomous exploration of unknown environments. \textit{TopoNav} dynamically constructs and maintains a topological map of the environment using a deep neural network that learns to extract task-relevant features from raw sensor observations. By combining the strengths of learning-based perception, hierarchical decision-making, and intrinsically motivated exploration, \textit{TopoNav} demonstrates significant improvements in efficiency, robustness, and adaptability compared to state-of-the-art navigation methods.

The key contributions of this work are:

\begin{enumerate}
\item \textbf{Enhanced Hierarchical Reinforcement Learning with Active Topological Mapping:} We introduce an enhanced hierarchical reinforcement learning framework that extends the Hierarchical Deep Q-Network (H-DQN) ~\cite{kulkarni2016hierarchical} architecture by integrating an actively updated topological map and leveraging intrinsic rewards to facilitate multi-level navigation policy learning. The meta-controller is responsible for choosing subgoals from the topological map, while the sub-controllers are designed to reach these subgoals through the execution of primitive actions. This dual strategy ensures that navigation is not only efficient but also directed towards regions of the environment that significantly enhance its understanding. By incorporating an intrinsically motivated learning approach, we effectively address the challenges associated with sparse extrinsic rewards, thereby accelerating the learning of efficient navigation policies.

\item \textbf{Dynamic Subgoal Generation and Strategic Landmark Selection:} We design a dynamic subgoal generation mechanism that activates upon detecting landmarks, trees, or objects while navigating. Detected features become part of the topological map as subgoals or nodes, facilitating structured navigation. When multiple landmarks are detected at similar distances, \textit{TopoNav} utilizes a strategic landmark selection strategy. This approach gives priority to landmarks that are most informative and relevant, considering their novelty and alignment with the final goal. This method promotes efficient exploration and ensures navigation is goal-oriented, even when presented with numerous potential subgoals.


\item \textbf{Experimental Validation of Superior Performance:} We extensively evaluate \textit{TopoNav} in diverse simulated environments and real-world scenarios, benchmarking against state-of-the-art baselines. It showcases a significant increase in exploration coverage (7-20\%), and navigation success rates (9-19\%), and achieves substantial reductions in navigation times (15-36\%) across various scenarios. These improvements demonstrate \textit{TopoNav}'s superior navigation in complex environments.

 
\end{enumerate}

The rest of the paper is organized as follows. Section~\ref{sec:related_work} discusses related work on robot navigation, topological mapping, and reinforcement learning. The background and problem formulation are discussed in Section~\ref{sec:background}. Section~\ref{sec:toponav_approach} presents the \textit{TopoNav} framework. Section~\ref{sec:experiments} describes the experimental setup, environments, and evaluation metrics. Section~\ref{sec:experiments} presents the results and analysis of the experiments, comparing \textit{TopoNav} with SOTA navigation methods. Finally, Section~\ref{sec:conclusion} concludes the paper and discusses future research directions.
\section{Related Work}
\label{sec:related_work}

This section discusses prior works related to autonomous robot navigation, focusing on learning-based methods, and topological mapping techniques.


\subsection{Learning-Based Navigation}
Recent advancements in deep reinforcement learning (RL) have led to the development of learning-based navigation methods that can learn effective policies directly from sensory inputs. Deep Q-Networks (DQN)~\cite{mnih2015human} have been applied to navigation tasks, enabling robots to learn collision-free paths in different environments. However, these methods typically require a large amount of training data and may struggle to generalize to unseen environments. Hierarchical reinforcement learning (HRL) methods have been proposed to address the challenges of sparse rewards and long-horizon tasks in navigation~\cite{kulkarni2016hierarchical, levy2019learning, li2019hierarchical}. These approaches typically learn a hierarchy of policies, where higher-level policies select subgoals or actions for lower-level policies to execute. However, most HRL methods ~\cite{li2020hrl4in} assume access to structured representations of the environment or rely on pre-defined subgoals, limiting their applicability in unknown and unstructured environments.

\subsection{Topological Mapping Navigation}
Topological mapping techniques represent the environment as a graph, where nodes correspond to distinct places and edges represent navigable paths between them~\cite{kuipers2000spatial, boal2014topological}. These methods focus on capturing the connectivity and adjacency information of the environment rather than maintaining a precise metric map. Spectral clustering algorithms have been used to construct topological maps from sensor data~\cite{liu2012robotic, blochliger2018topomap}. These methods exploit the eigenstructure of the similarity matrix to partition the environment into distinct regions. However, they often require a pre-specified number of clusters and may not adapt well to changes in the environment. Incremental topological mapping approaches incrementally build and refine the topological map as the robot explores the environment~\cite{kaess2008isam, ila2017slam++}. These methods typically use appearance-based place recognition techniques to detect loop closures and update the map accordingly. However, they may struggle in environments with significant changes in appearance. Recently, learning-based approaches have been proposed for topological mapping and localization~\cite{chaplot2020learning, chaplot2020neural}. These methods learn to extract topological representations directly from sensory inputs using deep neural networks. Suomela et al.~\cite{Suomela_2024_Placenav} proposed PlaceNav, a topological navigation approach that utilizes visual place recognition for subgoal selection. Wiyatno et al.~\cite{wiyatno2022lifelong} proposed a lifelong topological navigation approach that builds and continuously refines a sparse topological graph. However, they often require a large amount of training data and may not generalize well to unseen and sparse reward environments.

While existing topological navigation methods have shown promising results, they often rely on dense reward signals or extensive exploration to build effective representations of the environment. In contrast, \textit{TopoNav} is specifically designed to operate in sparse-reward settings by incorporating intrinsic motivation and hierarchical reinforcement learning, enabling efficient navigation and map construction with limited extrinsic feedback. 
By dynamically constructing and refining the topological map during exploration, \textit{TopoNav} can adapt to changes in the environment. The hierarchical policy architecture allows for effective navigation and obstacle avoidance, while intrinsic motivation guides exploration towards informative regions. 

\section{Background and Problem Formulation}

\label{sec:background}

In this section, we outline the topological mapping for navigation, the Hierarchical Deep Q-Networks (H-DQN) for policy learning, and our problem formulation.

\subsection{Topological Mapping}
Topological mapping represents the environment as a graph $\mathcal{G} = (\mathcal{V}, \mathcal{E})$, where nodes $v \in \mathcal{V}$ correspond to distinct places or landmarks, and edges $e \in \mathcal{E}$ represent the connectivity between them. This representation allows for efficient path planning and navigation by abstracting away the metric details and focusing on the high-level structure of the environment ~\cite{blochliger2018topomap}. In our approach, we dynamically construct and update the topological map $\mathcal{M}$ as the robot explores the environment. Each node $v_i \in \mathcal{V}$ is associated with a feature vector $\mathbf{f}_i$ that encodes the sensory observations at that location. The edges $e_{ij} \in \mathcal{E}$ represent the traversability between nodes $v_i$ and $v_j$, which can be determined based on the robot's motion model and the observed environmental conditions. The topological map $\mathcal{M}$ is used to guide the exploration process and enable efficient decision-making. By representing the environment as a graph, the robot can plan paths between different nodes and make high-level decisions based on the connectivity of the map. This allows for more strategic exploration and navigation compared to purely reactive approaches.

\subsection{Hierarchical Deep Q-Networks (H-DQN)}
Hierarchical Deep Q-Networks (H-DQN) ~\cite{kulkarni2016hierarchical} is a hierarchical reinforcement learning algorithm that extends the standard Deep Q-Networks (DQN) ~\cite{mnih2015human}. We choose the H-DQN due to its efficiency in managing tasks across different abstraction levels and its enhanced sample efficiency through off-policy learning and experience replay, which is crucial for real-world robotics applications where data collection can be costly and time-consuming. H-DQN learns a two-level policy, the high-level policy (meta-controller) learns to select subgoals from the topological map, while the low-level policies (sub-controllers) learn to generate actions to reach the selected subgoals. The meta-controller is represented by a Q-network $Q^{\mu}(s, g; \theta^{\mu})$, where $s \in \mathcal{S}$ is the current state, $g \in \mathcal{G}$ is a subgoal, and $\theta^{\mu}$ are the network parameters. The meta-controller selects subgoals based on the learned Q-values, which estimate the expected cumulative reward for reaching each subgoal. The sub-controllers are represented by a set of Q-networks ${Q^{g}(s, a; \theta^{g}) | g \in \mathcal{G}}$, where $s \in \mathcal{S}$ is the current state, $a \in \mathcal{A}$ is an action, and $\theta^{g}$ are the network parameters for subgoal $g$. Each sub-controller learns a policy to navigate from the current state to the corresponding subgoal by selecting actions that maximize the learned Q-values. The meta-controller and sub-controllers are trained simultaneously with separate replay buffers and target networks for each level of the hierarchy.

\subsection{Problem Formulation}
We formulate the problem as a Markov Decision Process (MDP) with a hierarchical structure. The MDP is defined by a tuple $(\mathcal{S}, \mathcal{A}, \mathcal{P}, \mathcal{R}, \gamma)$, where $\mathcal{S}$ is the state space, $\mathcal{A}$ is the action space, $\mathcal{P}$ is the transition probability function, $\mathcal{R}$ is the reward function, and $\gamma$ is the discount factor. The state space $\mathcal{S}$ consists of the robot's sensory observations and the current topological map $\mathcal{M}$. The action space $\mathcal{A}$ is hierarchically structured, with high-level actions (subgoals) $\mathcal{A}_h$ selected by the meta-controller and low-level actions (primitives) $\mathcal{A}_l$ executed by the controller. The transition probability function $\mathcal{P}: \mathcal{S} \times \mathcal{A} \rightarrow \mathcal{S}$ determines the next state based on the current state and the executed action. The reward function $\mathcal{R}: \mathcal{S} \times \mathcal{A} \rightarrow \mathbb{R}$ is sparse, providing positive rewards for reaching the goal or completing milestones, and intrinsic rewards for exploring new nodes or discovering new connections in the topological map. The objective is to learn a hierarchical policy $\pi = (\pi_h, \pi_l)$ that maximizes the expected cumulative reward:

\begin{equation}
\pi^* = \argmax_{\pi} \mathbb{E}_{\pi} \left[ \sum_{t=0}^{\infty} \gamma^t r_t \right],
\end{equation}
where $\pi^*$ is the optimal policy, $\pi_h: \mathcal{S} \rightarrow \mathcal{A}_h$ is the high-level policy (meta-controller), $\pi_l: \mathcal{S} \times \mathcal{A}_h \rightarrow \mathcal{A}_l$ is the low-level policy (controller), $r_t$ is the reward at time step $t$, and $\gamma$ is the discount factor. In this context, the state space $\mathcal{S}$ includes a dynamically constructed topological map $\mathcal{M}$, where nodes represent automatically identified landmarks. The high-level action space $\mathcal{A}_h$ consists of subgoals, allowing the meta-controller to make decisions based on the evolving map. This approach enables efficient navigation in unknown environments without relying on predefined waypoints.

\section{TopoNav: Topological Map Generation and Navigation}
\label{sec:toponav_approach}
\begin{figure}[!htb]
    \centering
    \includegraphics[width=0.5\textwidth]{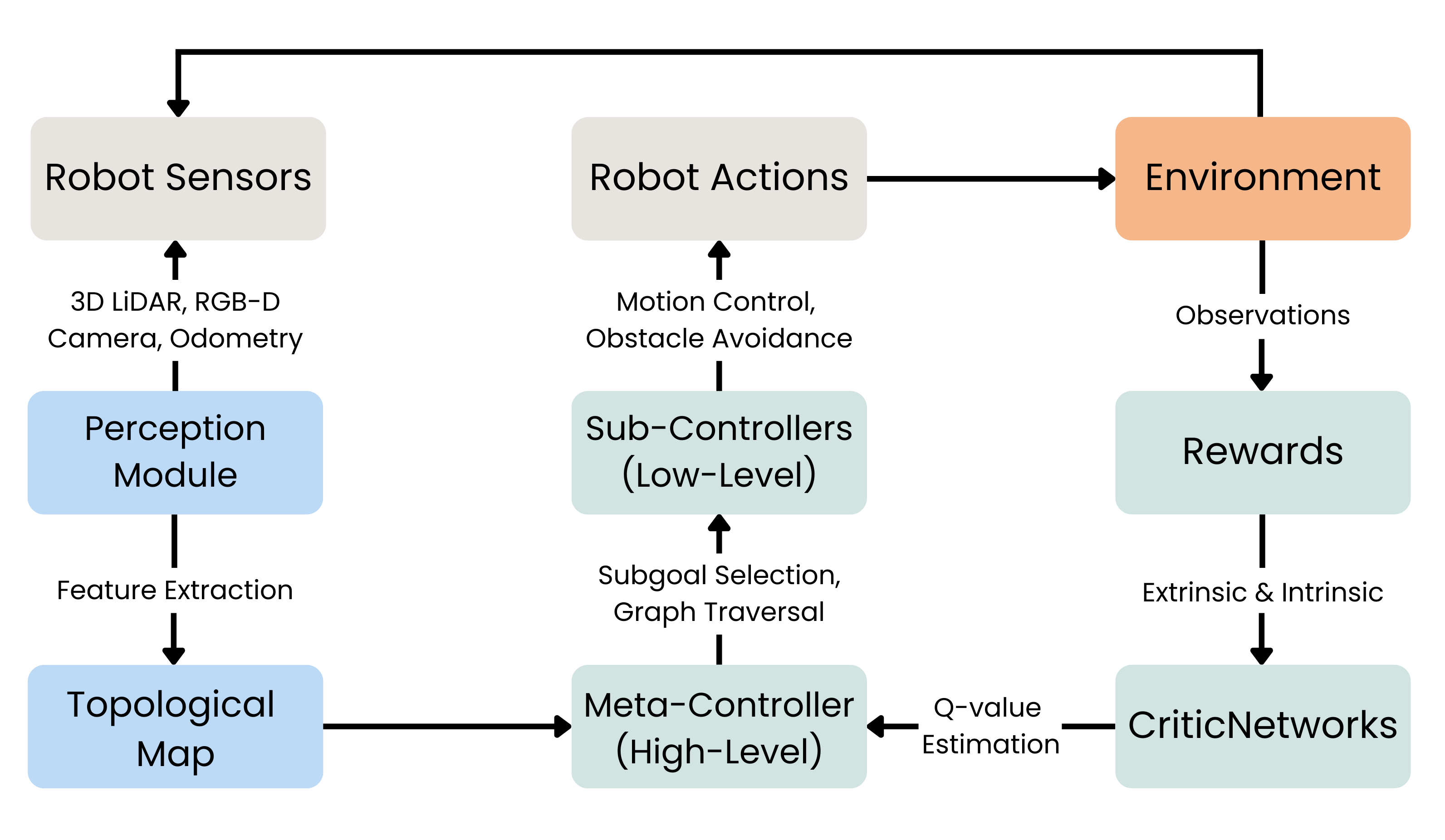}
    \caption{Overview of \textit{TopoNav} System Architecture.}     
    \label{fig:architecture}
\end{figure}

In this section, We explain the major stages of the \textit{TopoNav} approach. Fig. \ref{fig:architecture} shows how different modules in our method are connected. 

\subsection{Attention-based Feature Detection}
\textit{TopoNav} integrates an attention-based feature detection module, combining a ResNet-50 CNN~\cite{he2016deep} with a Convolutional Block Attention Module (CBAM)~\cite{woo2018cbam}, to detect landmarks, objects, and trees from RGB images $\mathbf{I}_i \in \mathbb{R}^{H \times W \times 3}$, where $H$ and $W$ are the height and width of the input image, respectively. This module processes images to produce feature maps $\mathbf{F}_i \in \mathbb{R}^{h \times w \times c}$ ($h$ and $w$ are the height and width of the feature map, and $c$ is the number of channels) refining them through CBAM to emphasize relevant features for landmark detection. Utilizing a sliding window approach on the refined feature map $\mathbf{F}'_i$, \textit{TopoNav} identifies potential Regions of Interest (ROIs) corresponding to navigational markers. Each ROI's feature vector $\mathbf{f}_{ROI} \in \mathbb{R}^d$, obtained through ROI pooling, undergoes a binary classification to ascertain the presence of valid landmarks, objects, or trees, assigning a probability score $\mathbf{P}_{ROI} \in [0, 1]$. For each detected landmark, object, or tree, \textit{TopoNav} extracts its corresponding feature vector $\mathbf{f}_{ROI}$ and uses it to create a new node or subgoal in the topological map. 

\subsection{Subgoal Generation and Navigation}
\label{subsec:subgoal_generation_navigation}

In \textit{TopoNav}, subgoals or nodes are generated dynamically based on the landmarks, trees, or objects encountered by the robot while navigating towards the final goal. Whenever a new landmark or object is detected, it is compared with the existing nodes in the topological map using a similarity measure (described in Section~\ref{subsec:reward_structure_exploration}). If the similarity is below a threshold, the landmark is added as a new subgoal or node to the topological map. However, when multiple landmarks are detected at similar distances, \textit{TopoNav} employs a selection strategy to prioritize the most informative and relevant landmark as the next subgoal based on novelty and goal-directedness. The novelty of a landmark $l$ is calculated based on the number of previous visits, using a novelty factor that decreases exponentially with the number of visits:

\begin{equation}
N(l) = e^{-\lambda \cdot \text{visits}(l)}
\end{equation}

where $\lambda$ is a decay parameter controlling the rate at which the novelty factor decreases with visits. The goal-directedness of a landmark is calculated as the cosine similarity between the direction vector $\mathbf{v}_l$ towards the landmark $l$ and the direction vector towards the goal $\mathbf{v}_g$. The landmark with the highest score is selected as the next subgoal. The weights $w_N$ and $w_{GD}$ adjust the importance of novelty and goal-directedness in landmark selection, balancing exploration, and direct goal attainment. (See Algorithm ~\ref{alg:landmark_selection})

\begin{equation}
GD(l) = \frac{\mathbf{v}_l \cdot \mathbf{v}_g}{||\mathbf{v}_l|| \cdot ||\mathbf{v}_g||}
\end{equation}


\begin{algorithm}
\caption{Strategic Landmark Selection}
\label{alg:landmark_selection}
\DontPrintSemicolon
\KwIn{Detected landmarks $L$, current position $\mathbf{p}$, goal position $\mathbf{g}$}
\KwOut{Selected landmark $l_{\text{best}}$}

Initialize $\text{max\_score} = -\infty$\;
\ForEach{landmark $l \in L$}{
    $N(l) = e^{-\lambda \cdot \text{visits}(l)}$\;
    $\mathbf{v}_l = \text{direction\_vector}(\mathbf{p}, l)$\;
    $\mathbf{v}_g = \text{direction\_vector}(\mathbf{p}, \mathbf{g})$\;
    $GD(l) = \frac{\mathbf{v}_l \cdot \mathbf{v}_g}{||\mathbf{v}_l|| \cdot ||\mathbf{v}_g||}$\;
    $\text{score}(l) = w_N \cdot N(l) + w_{GD} \cdot GD(l)$\;
    \If{$\text{score}(l) > \text{max\_score}$}{
        $\text{max\_score} = \text{score}(l)$\;
        $l_{\text{best}} = l$\;
    }
}
\Return $l_{\text{best}}$\;
\end{algorithm}

\begin{figure}[!htb]
    \centering
    \resizebox{\columnwidth}{!}
    { 
        \begin{tikzpicture}[
            node_style/.style={circle, draw=blue!50, fill=blue!20, thick, minimum size=10mm},
            landmark_style/.style={circle, draw=orange!50, fill=orange!20, thick, minimum size=8mm},
            edge_style/.style={draw=black, thick, ->},
            dashed_edge/.style={draw=black, thick, dashed, ->}
        ]

        \node[node_style] (X_0) at (0, 0) {Start};

        \node[landmark_style] (N_01) at (2, 2) {$N_{01}$};
        \node[right] at (N_01.east) {Sub-goal};

        \node[landmark_style] (N_12) at (4, -1) {$N_{12}$};
        \node[right] at (N_12.east) {Sub-goal};

        \node[landmark_style] (N_23) at (6, 2) {$N_{23}$};
        \node[right] at (N_23.east) {Sub-goal};

        \node[node_style] (X_goal) at (8, 0) {Goal};

        \draw[edge_style] (X_0) -- (N_01) node[midway, right, xshift=5pt] {Navigate};
        \draw[edge_style] (N_01) -- (N_12) node[midway, right, xshift=8pt, yshift=-5pt] {Navigate};
        \draw[edge_style] (N_12) -- (N_23) node[midway, right, xshift=10pt, yshift=5pt] {Navigate};
        \draw[edge_style] (N_23) -- (X_goal) node[midway, right, xshift=10pt] {Navigate};

        \draw[dashed_edge] (N_01) -- ++(-1,1) node[above] {Select $l_{best}$ using $N(l), GD(l)$};
        \draw[dashed_edge] (N_12) -- ++(0,-1.5) node[below, yshift=-5pt] {Select $l_{best}$ using $N(l), GD(l)$};
        \draw[dashed_edge] (N_23) -- ++(1,1) node[above] {Select $l_{best}$ using $N(l), GD(l)$};

        \node at (-2, 1.3) {Meta-Controller/Sub-Controller};
        \draw[dashed_edge] (-1, 1) -- (X_0);

        \end{tikzpicture}
    }
   \caption{An illustration of the topological navigation process using detected landmarks as sub-goals and strategic landmark selection.}
    \label{fig:topo_navigation}
\end{figure}
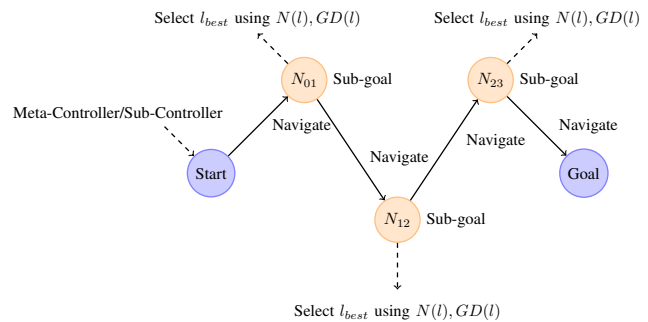

\begin{algorithm}
\caption{TopoNav: Topological Map Generation and Navigation}
\label{alg:toponav}
\DontPrintSemicolon 
\SetAlgoNlRelativeSize{-1}
\SetNlSty{textbf}{}{:}
\SetAlgoNlRelativeSize{-2}
\KwIn{$G_g$, the global end goal; $d_{\text{thresh}}$, distance threshold for subgoal generation}
\KwOut{$T_m$, the topological map with nodes ($N$) and edges ($E$)}
\BlankLine
Initialize meta-controller $Q^\mu$ and sub-controllers $Q^g$.\;
Initialize replay buffers $\mathcal{D}^\mu, \mathcal{D}^g$ for each subgoal $g$.\;
$T_m \leftarrow \emptyset$; $s_0 \leftarrow \text{GetInitialState()}$; $g_0 \leftarrow G_g$.\;
\While{$g_t \neq G_g$}{
    $\mathcal{L} \leftarrow \text{DetectLandmarks}(s_t)$.\;
    \eIf{$\text{len}(\mathcal{L}) > 0$}{
        $l_{\text{best}} \leftarrow \text{Use Algorithm ~\ref{alg:landmark_selection}}(\mathcal{L}, s_t, G_g)$.\;
        \If{$\text{IsNewNode}(l_{\text{best}})$}{
            Add $l_{\text{best}}$ to $T_m$; $g_t \leftarrow l_{\text{best}}$.\;
        }
    }{
        \If{$\text{dist}(s_t, g_t) > d_{\text{thresh}}$}{
            $g' \leftarrow$ Generate new subgoal; Add $g'$ to $T_m$; $g_t \leftarrow g'$.\;
        }
    }
    \eIf{$g_t \neq G_g$}{
        $a_t \leftarrow$ Select action; Execute $a_t$; Observe $s_{t+1}, r_t$; Store transition.\;
        \If{reached $g_t$}{
            $g_t \leftarrow G_g$.\;
        }
    }{
        Update subgoal; Store and update transition.\;
    }
    $s_t \leftarrow s_{t+1}$.\;
}
\Return{$T_m$}\;
\end{algorithm}

During navigation, the meta-controller selects the next subgoal based on the current state and the robot's overall goal. If no landmarks are detected within a certain distance threshold, \textit{TopoNav} generates a new subgoal along the robot's current trajectory at the maximum detection range to ensure continuous progress towards the final goal. The sub-controller's policy generates a sequence of primitive actions or low-level skills to navigate the robot towards the selected subgoal, aiming to maximize the expected cumulative reward. The navigation process continues until the robot reaches the final goal or a maximum number of steps is exceeded. (See Algorithm ~\ref{alg:toponav} \& Fig. \ref{fig:topo_navigation})


\subsection{Reward Structure}
\label{subsec:reward_structure_exploration}
To address the challenges of sparse reward environments, \textit{TopoNav} employs a carefully designed reward structure that combines sparse extrinsic rewards, dense intrinsic rewards, and penalties for suboptimal exploration behavior. The total reward $r$ received by the agent at each time step is calculated as follows:
\begin{align}
r = & \, \alpha \cdot r_{ex} + \beta \cdot (r_{in} + r_{sg} + r_{fe} + r_{ep} + r_{ue}) \nonumber \\
    & + \gamma \cdot (r_{p} + r_{sd} + r_{te} + r_{ob})
\end{align}
where $r_{ex}$ represents the sparse extrinsic reward, which is given only when the agent reaches the final goal ($R_{goal}$) or completes a milestone ($R_{milestone}$):

\begin{equation}
r_{ex} =
\begin{cases}
R_{goal}, & \text{if } s_{t+1} \text{ is the final goal} \\
R_{milestone}, & \text{if } s_{t+1} \text{ completes a milestone} \\
0, & \text{otherwise}
\end{cases}
\end{equation}

The intrinsic rewards $r_{in}$, $r_{sg}$, $r_{fe}$, $r_{ep}$, and $r_{ue}$ encourage exploration and map expansion. $r_{in}$ is based on the novelty of the visited states, calculated as the inverse square root of the visitation count:
\begin{equation}
r_{in} = \frac{1}{\sqrt{N(s_t)}}
\end{equation}

$r_{sg}$ is a constant reward given when the agent discovers a new subgoal. The frontier exploration reward $r_{fe}$ encourages the agent to prioritize the exploration of frontier nodes in the topological map:
\begin{equation}
r_{fe} = \lambda_{fe} \cdot \frac{N_{fn}}{N_{tn}}
\end{equation}

where $N_{fn}$ is the number of new frontier nodes reached, $N_{tn}$ is the total number of nodes in the topological map, and $\lambda_{fe}$ is a scaling factor. The exploration progress reward $r_{ep}$ rewards the agent for increasing the explored area or adding new nodes to the topological map:

\begin{equation}
r_{ep} = \lambda_{ep} \cdot \frac{\Delta A}{A_{total}}
\end{equation}

where $\Delta A$ is the increase in the explored area, $A_{total}$ is the total area of the environment, and $\lambda_{ep}$ is a scaling factor. The uncertainty-driven exploration reward $r_{ue}$ encourages the agent to explore regions where its knowledge is limited:

\begin{equation}
r_{ue} = \lambda_{ue} \cdot U(s)
\end{equation}

where $U(s)$ is the uncertainty estimate of the agent's knowledge about the current state $s$, and $\lambda_{ue}$ is a scaling factor. 
To discourage suboptimal exploration behavior, \textit{TopoNav} incorporates several penalty terms. The penalty for revisiting states $r_p$ discourages the agent from excessively revisiting previously explored areas:

\begin{equation}
r_p = -\lambda_p \cdot \frac{N(s_t) - 1}{\sqrt{N(s_t)}}
\end{equation}

where $N(s_t)$ is the number of times the state $s$ has been visited, and $\lambda$ is a scaling factor that controls the magnitude of the penalty. The term $\frac{N(s_t) - 1}{\sqrt{N(s_t)}}$ ensures that the penalty increases as the number of revisits grows, but with a diminishing effect to avoid completely discouraging revisits. The subgoal diversity penalty $r_{sd}$ penalizes the agent for selecting subgoals that are similar to previously visited subgoals:

\begin{equation}
r_{sd} = -\lambda_{sd} \cdot \max\limits_{g_h \in \mathcal{H}} \text{sim}(g_t, g_h)
\end{equation}

where $g_t$ is the current subgoal, $\mathcal{H}$ is the history of visited subgoals, $\text{sim}(\cdot)$ is a similarity function, and $\lambda_{sd}$ is a scaling factor. In \textit{TopoNav}, the similarity function $\text{sim}(\cdot)$ is based on the Euclidean distance between the feature vectors associated with the subgoals. Each subgoal $g$ is represented by a feature vector $f_g$ that encodes its visual and spatial characteristics. The similarity between two subgoals $g_t$ and $g_h$ is calculated as:

\begin{equation}
\text{sim}(g_t, g_h) = \sqrt{\sum_{i=1}^{n} (f_{g_t}^{(i)} - f_{g_h}^{(i)})^2}
\label{eq:similarity}
\end{equation}

This similarity measure allows \textit{TopoNav} to quantify the dissimilarity between subgoals based on their visual and spatial attributes, promoting the selection of diverse subgoals during exploration. The temporal exploration penalty $r_{te}$ penalizes the agent for long periods of non-exploration:

\begin{equation}
r_{te} = -\lambda_{te} \cdot (t - t_{last_exp})
\end{equation}

where $t$ is the current time step, $t_{last_exp}$ is the time step of the last exploration progress, and $\lambda_{te}$ is a scaling factor. An additional penalty $r_{ob}$ is introduced to penalize the agent for hitting obstacles:

\begin{equation}
r_{ob} =
\begin{cases}
R_{obstacle}, & \text{if the agent hits an obstacle} \\
0, & \text{otherwise}
\end{cases}
\end{equation}

where $R_{obstacle}$ is a negative constant reward. This penalty encourages the agent to avoid collisions and navigate safely in the environment.
The hyperparameters $\alpha$, $\beta$, and $\gamma$ balance the contributions of the extrinsic rewards, intrinsic rewards, and penalties, respectively. 

\subsection{Hierarchical Policy Learning}
\label{subsec:hierarchical_policy_learning}

\textit{TopoNav} utilizes the H-DQN algorithm to develop navigation policies across different abstraction levels. This framework employs a meta-controller for high-level subgoal selection and sub-controllers for detailed navigation tasks.
The meta-controller, leveraging an attention-based feature vector $\mathbf{f}_t$ from current observations, selects subgoals from the topological map using a DQN-based policy $\pi_m(s_m, g; \theta_m)$. This selection process prioritizes key locations or landmarks in the environment, informed by both the robot's current state and the desired goal features. On the other hand, sub-controllers focus on reaching these subgoals by executing actions derived from a separate DQN policy $\pi_s(s_s, a; \theta_s)$, guided by the local obstacle map and the robot's kinematics to ensure safe navigation. The carefully designed reward structure (described in section ~\ref{subsec:reward_structure_exploration}) plays an important role in this hierarchy. Training alternates between meta and sub-controllers using their respective experiences and policies, optimizing a composite reward structure that balances exploration, safety, and goal orientation. This approach enables \textit{TopoNav} to adaptively learn efficient navigation strategies in sparse-reward settings, leveraging the strengths of hierarchical learning and attention-based feature extraction.



\textbf{Theorem 1 (Subgoal Reachability)}. \textit{Given a topological map $G = (V, E)$ constructed by TopoNav, with $V$ representing nodes (subgoals) and $E$ representing edges (navigable paths), for any two nodes $v_i, v_j \in V$, if a path exists from $v_i$ to $v_j$ in $G$, then the sub-controller policy $\pi_s$ learned by TopoNav can navigate the robot from $v_i$ to $v_j$ with a probability of at least $1 - \epsilon$. Here, $\epsilon$ is a small positive constant representing the upper bound of navigation failure probability, which accounts for environmental stochasticity and the convergence of the learning algorithm. TopoNav minimizes $\epsilon$ through robust feature extraction and continuous map refinement.}



\textbf{Proof}. Consider a path $P_{ij} = \{v_i, e_{i1}, v_1, ..., e_{(k-1)j}, v_j\}$ between nodes $v_i$ and $v_j$ in the topological map $G$, incorporating both nodes and the edges $e_{pq} \in E$ between them. The sub-controller policy $\pi_s$, leveraging DQN methodologies, aims for a navigation success across each edge $e_{pq}$ with a probability $\geq 1 - \epsilon_{pq}$, where $\epsilon_{pq}$ signifies the maximum probability of failing to navigate from $v_p$ to $v_q$ successfully. The overall probability of navigating the entire path $P_{ij}$, as the product of the success probabilities for its constituting edges, is $\prod_{e_{pq} \in P_{ij}} (1 - \epsilon_{pq})$. This is theoretically at least $1 - \epsilon$, where $\epsilon = k \max_{p,q} \epsilon_{pq}$ represents the compounded probability of any failure occurring along the path with $k$ edges. Therefore, with the sub-controller policy $\pi_s$, \textit{TopoNav} can proficiently facilitate navigation from $v_i$ to $v_j$ within the given environmental conditions and learning algorithm constraints, maintaining a success probability of at least $1 - \epsilon$. $\square$

\section{Experiments and Results}
\label{sec:experiments}
In this section, we present the experimental setup, results, and analysis of the \textit{TopoNav} framework. We evaluate the performance of \textit{TopoNav} in both simulated and real-world environments and compare it with state-of-the-art baselines.

\subsection{Evaluation Environments}
The \textit{TopoNav} framework is evaluated in a range of simulated and real-world environments. The simulated environments are created using the Unity 3D engine and include outdoor urban scenes and off-road terrains. The simulated environments incorporate landmarks and objects similar to those in our real-world test scenarios (Fig. \ref{fig:simulation_unity}). The real-world experiments are conducted in an outdoor off-road environment using a Clearpath Jackal robot. The environments are designed to cover various navigation challenges, such as narrow passages, obstacles, uneven terrain, and dead ends. The size of the environments ranges from 20m × 20m to 200m × 200m, and the complexity is varied by adjusting the density and arrangement of obstacles. For each environment, multiple starting and goal locations are randomly sampled to create a diverse set of navigation tasks. The performance of \textit{TopoNav} is evaluated over 100 episodes in each environment, and the average metrics are reported.

\subsection{Evaluation Metrics}
We evaluate the performance of \textit{TopoNav} and the baselines using the following metrics:
\begin{itemize}
\item \textbf{Success Rate}: The percentage of successful navigation episodes where the robot reaches the goal location within a specified time limit.
\item \textbf{Navigation Time}: The average time taken by the robot to reach the goal location in successful episodes.
\item \textbf{Trajectory Length}: The average length of the robot's trajectory in successful episodes.
\item \textbf{Exploration Coverage}: The percentage of the environment explored by the robot during navigation.
\end{itemize}

\subsection{Baselines}
To rigorously evaluate \textit{TopoNav}'s performance, we benchmark it against a diverse set of state-of-the-art baselines:
\begin{itemize}
    

    \item \textbf{PlaceNav}: A topological navigation approach that utilizes visual place recognition for subgoal selection and integrates a Bayesian filter to improve the temporal consistency of subgoal selection \cite{Suomela_2024_Placenav }.
    
    \item \textbf{TopoMap}: A method focusing on navigation through the construction of topological maps from predefined landmarks \cite{blochliger2018topomap}, providing a direct comparison to assess the benefits of \textit{TopoNav}'s dynamic mapping and navigation strategy.

    \item \textbf{ViNG (Visual Navigation with Goals)}: A learning-based navigation approach that learns to navigate to visual goals in open-world environments by combining hierarchical planning and topological mapping \cite{shah2021ving}.

    \item \textbf{Lifelong Topological Visual Navigation (LTVN)}: A topological navigation method that builds and continuously refines a sparse topological graph \cite{wiyatno2022lifelong}.
\end{itemize}

\begin{figure*}[!htb]
    \centering
    \subfloat[]{%
        \includegraphics[width=0.24\textwidth,height=3cm]{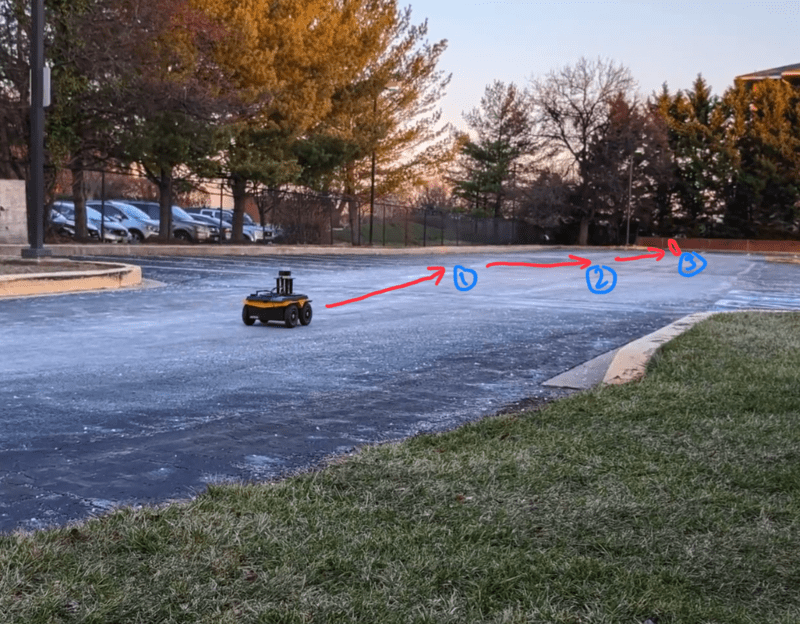}
    }
    \hfill 
    \subfloat[]{%
        \includegraphics[width=0.24\textwidth,height=3cm]{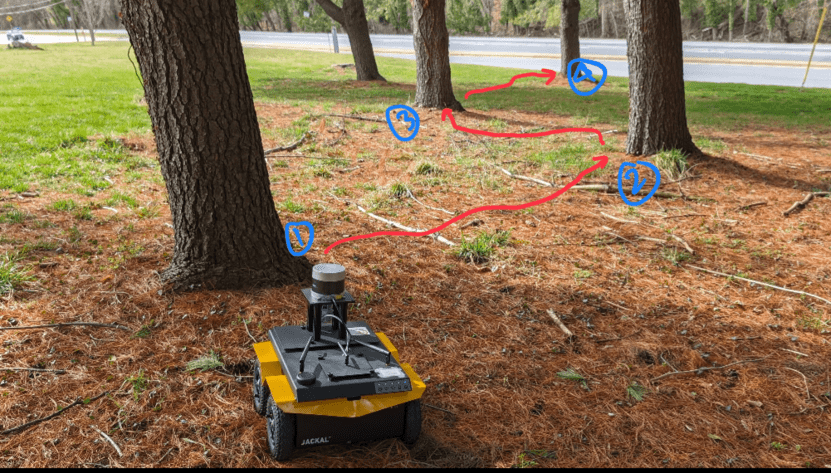}
    }
    \hfill
    \subfloat[]{%
        \includegraphics[width=0.24\textwidth,height=3cm]{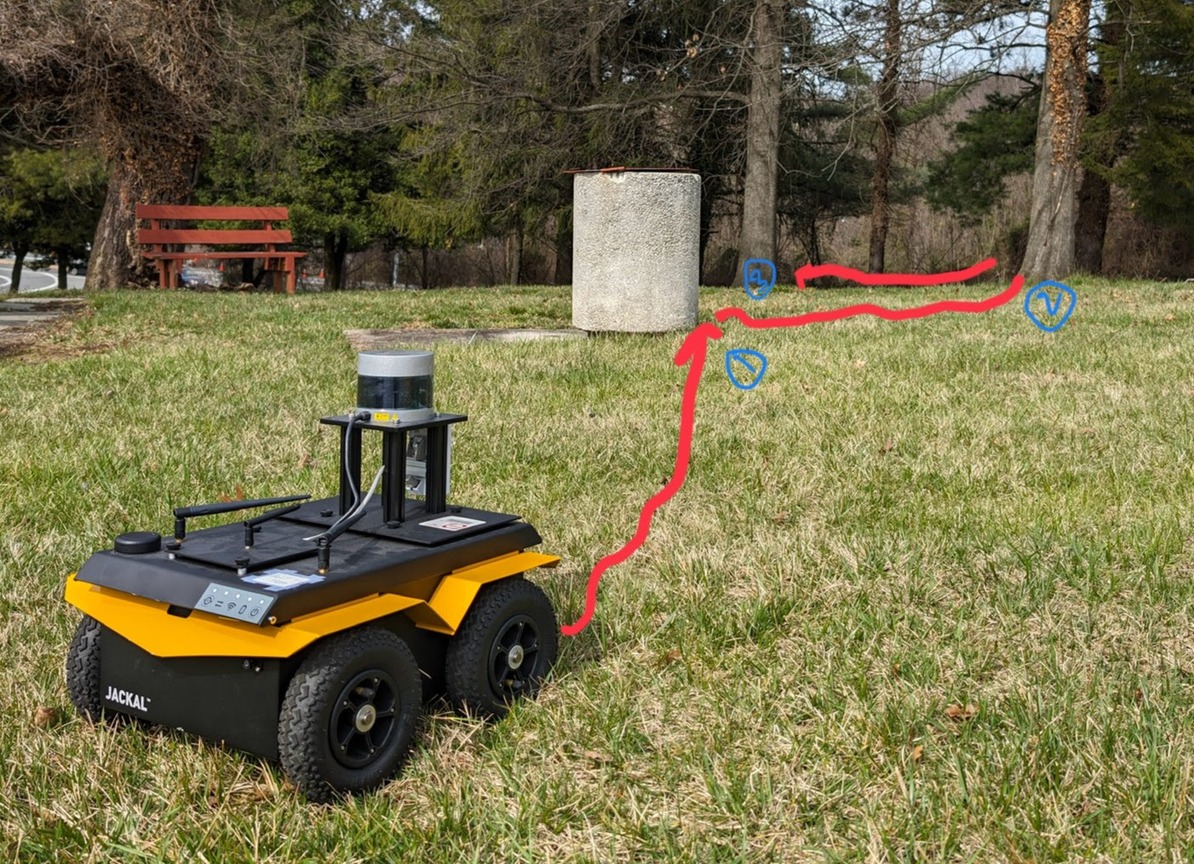}
    }
    \hfill
    \subfloat[]{%
        \includegraphics[width=0.24\textwidth,height=3cm]{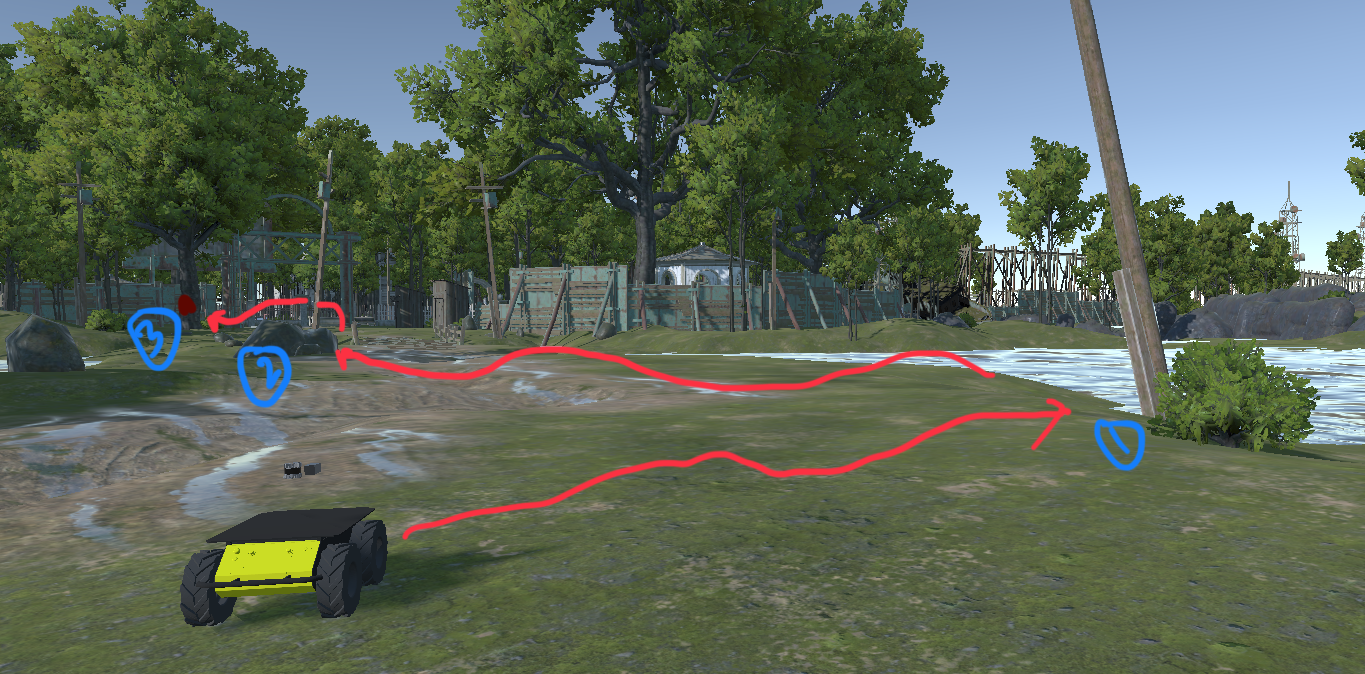}
        \label{fig:simulation_unity}
    }

    \subfloat[]{%
        \includegraphics[width=0.24\textwidth,height=3cm]{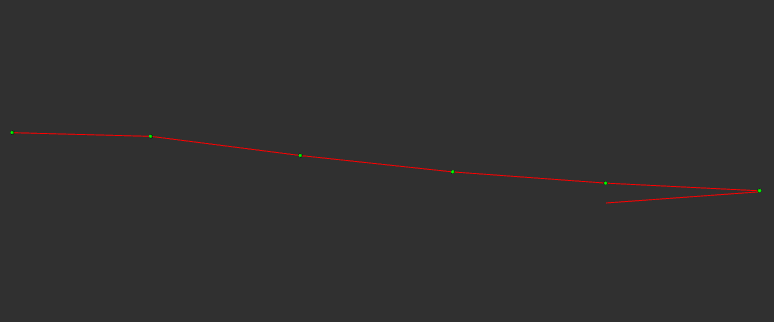}
    }
    \hfill
    \subfloat[]{%
        \includegraphics[width=0.24\textwidth,height=3cm]{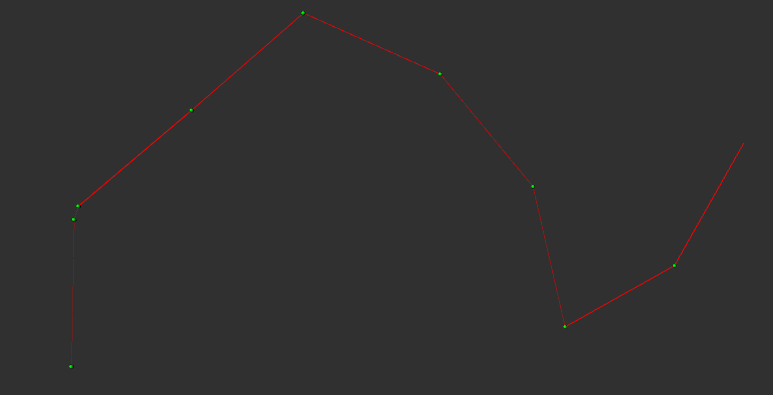}
    }
    \hfill
    \subfloat[]{%
        \includegraphics[width=0.24\textwidth,height=3cm]{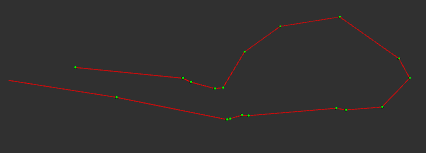}
    }
    \hfill
    \subfloat[]{%
        \includegraphics[width=0.24\textwidth,height=3cm]{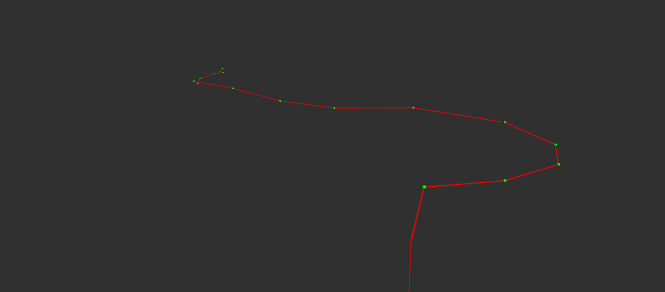}
    }

    \caption{ (a-d) The robot navigates through various outdoor environments: (a) an open space without obstacles/vegetation (Scenario 1), (b) a natural setting with trees/vegetation (Scenario 2), (c) a cluttered environment with obstacles and landmarks (Scenario 3), and (d) a diverse terrain in simulation. (e-h) The corresponding topological maps generated by \textit{TopoNav} for each scenario. The topological maps capture the connectivity and traversability of the environments, representing key locations and paths. The green dots in the maps represent the nodes or subgoals, which correspond to landmarks or distinct places. The edges (red lines) indicate the navigability between these nodes, enabling efficient path planning and navigation for the robot in each setting.}
    \label{fig:environments_maps}
\end{figure*}


\begin{figure}[!htb]
    \centering
    \subfloat[\centering]{
        \includegraphics[width=.46\columnwidth,height=3.5cm]{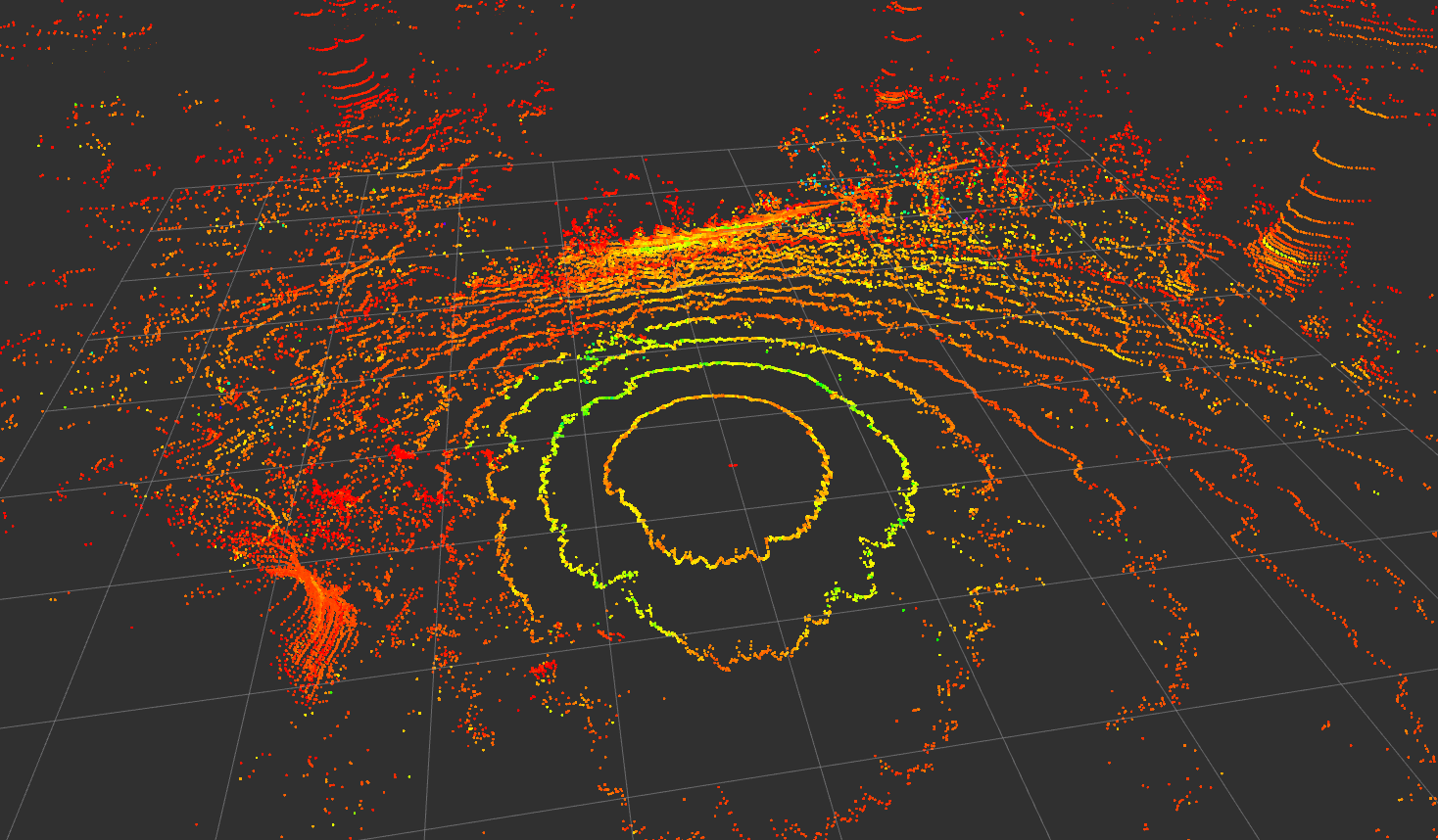}
    }
    \hspace{-0.5em} 
    \subfloat[\centering]{
        \includegraphics[width=.46\columnwidth,height=3.5cm]{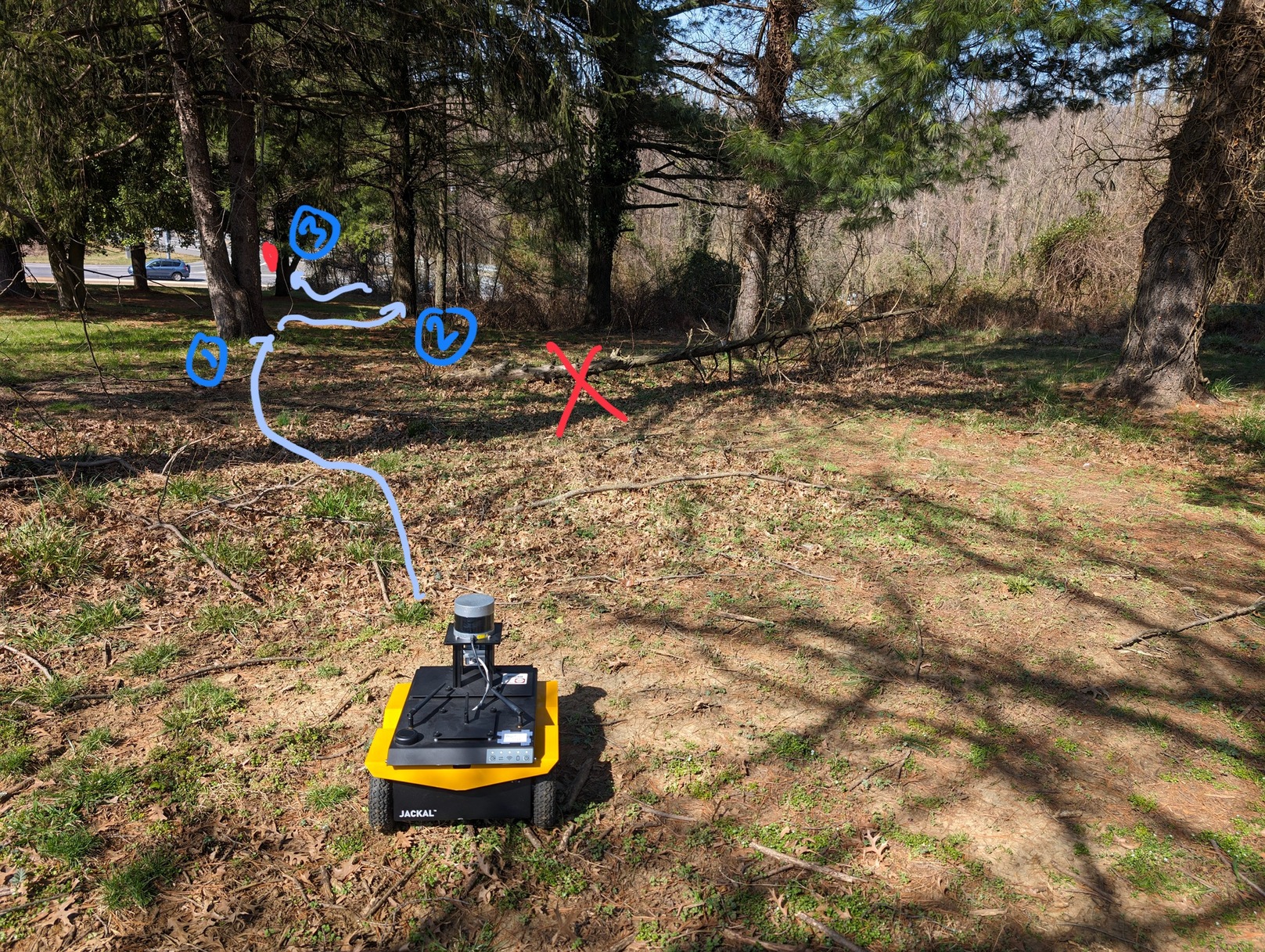}
    }
    \caption{
        (a) A 3D point cloud representation of the outdoor environment obtained from the robot's LiDAR sensor, showcasing the terrain and obstacle details.
        (b) Robot navigating through a physical environment, guided by the \textit{TopoNav} framework, executes an immediate obstacle avoidance maneuver to select an alternative route when faced with an obstruction (red cross, fallen trees).
    }
    \label{fig:obstacle_experiment}
\end{figure}

\begin{table}[ht]
\centering
\scriptsize
\begin{tabular}{|l|c|c|c|c|}
\hline
\textbf{Metrics} & \textbf{Methods} & \textbf{Scenario 1} & \textbf{Scenario 2} & \textbf{Scenario 3} \\
\hline
\textbf{Success} & PlaceNav & 89 & 87 & 88 \\
\textbf{Rate (\%)}& TopoMap & 82 & 79 & 80 \\
& ViNG & 84 & 82 & 83 \\
& LTVN & 90 & 88 & 89 \\
& \textbf{TopoNav (Ours)} & \textbf{98} & \textbf{94} & \textbf{92} \\
\hline
\textbf{Navigation} & PlaceNav & 37.5 & 38.2 & 39.0 \\
\textbf{Time (s)} & TopoMap & 47.5 & 46.9 & 48.2 \\
& ViNG & 39.8 & 40.5 & 41.2 \\
& LTVN & 36.2 & 37.0 & 37.8 \\
& \textbf{TopoNav (Ours)} & \textbf{30.6} & \textbf{33.8} & \textbf{35.1} \\
\hline
\textbf{Trajectory} & PlaceNav & 13.8 & 14.2 & 14.5 \\
\textbf{Length (m)} & TopoMap & 15.8 & 15.2 & 15.7 \\
& ViNG & 14.4 & 14.9 & 15.3 \\
& LTVN & 12.9 & 13.5 & 13.8 \\
& \textbf{TopoNav (Ours)} & \textbf{10.4} & \textbf{11.6} & \textbf{12.3} \\
\hline
\textbf{Exploration} & PlaceNav & 83 & 81 & 82 \\
\textbf{Coverage (\%)} & TopoMap & 75 & 73 & 74 \\
& ViNG & 79 & 77 & 78 \\
& LTVN & 84 & 83 & 85 \\
& \textbf{TopoNav (Ours)} & \textbf{90} & \textbf{88} & \textbf{89} \\
\hline
\end{tabular}
\caption{Comparative Analysis: \textit{TopoNav} vs. SOTA methods across scenarios, highlighting performance in three distinct environments (Section~\ref{sec:testingscenario}).}
\label{tab:real_results}
\end{table}

\subsection{Implementation Details}
Our RL network is implemented in PyTorch and trained using simulated terrains with a Clearpath Husky robot in ROS Noetic and Unity Simulation framework, which allows for testing in realistic outdoor environments with diverse terrain and objects.
The simulated Husky robot is mounted with a Velodyne VLP16 3D LiDAR. The network is trained in a workstation with a 10th-generation Intel Core i9-10850K processor and an NVIDIA GeForce RTX 3090 GPU. During training, our network uses a batch size of 128 and performs gradient updates using the Adam optimizer with learning rate $\lambda = 10^{-4}$. The training process occurs over 1 million steps in our simulation environment, with performance evaluated at regular intervals. For real-time deployment and inference, we use the Jackal UGV from Clearpath Robotics equipped with a 3D VLP-32C Velodyne Ultrapuck LiDAR and an AXIS Fixed IP Camera. It also includes an onboard Intel computer system, equipped with an Intel i7-9700TE CPU and an NVIDIA GeForce GTX 1650 Ti GPU.  We utilize the LiDAR sensor provided 3D point cloud data for obstacle detection and avoidance using the Dynamic Window Approach (DWA) ~\cite{fox1997dynamic}.

\subsection{Testing Scenario}
\label{sec:testingscenario}
We evaluate our topology based navigation framework in three scenarios (see Fig. \ref{fig:environments_maps}). 

\begin{itemize}[leftmargin=*]

\item \textbf{Scenario 1 (Goal Reaching):} In this scenario, the robot is placed in an obstacle-free outdoor environment with an objective of reaching a given goal location.

\item \textbf{Scenario 2 (Feature-Based Navigation):} In this scenario, the robot is placed in an outdoor area with static natural features such as trees, rocks, structures, and sharp hills. The goal is to reach the destination using the natural features for localization and mapping, challenging its environmental perception and landmark utilization.

 \item \textbf{Scenario 3 (Navigating on Complex Terrains):} In this scenario, the robot is placed in an environment with both obstacles and landmarks, this scenario assesses the robot's simultaneous obstacle avoidance and feature-based navigation, testing its adaptability in complex settings.
 \end{itemize}
 
\subsection{Performance Comparison and Analysis}
Table~\ref{tab:real_results} demonstrates \textit{TopoNav}'s significant improvements over existing navigation systems. Our framework outperforms baseline methods across key performance metrics, highlighting its effectiveness in real-world scenarios.
In goal-reaching tasks, \textit{TopoNav} achieves a 98\% success rate, surpassing PlaceNav~\cite{Suomela_2024_Placenav}, LTVN~\cite{wiyatno2022lifelong}, and ViNG~\cite{shah2021ving}. These baselines often struggle with efficient path planning in sparsely rewarded environments. In contrast, \textit{TopoNav} leverages its hierarchical structure and dynamic topological mapping for more effective route optimization. The limitations of the baselines become evident in feature-based navigation, where they fail to consistently utilize natural landmarks for navigation. \textit{TopoNav} excels in this scenario, achieving an 94\% success rate. This demonstrates its superior ability to accurately navigate by effectively using environmental features for guidance. The complex terrain scenario further highlights \textit{TopoNav}'s robustness. With a 92\% success rate, it significantly outperforms baseline systems, which often experience performance degradation in challenging navigational conditions. \textit{TopoNav}'s adaptability to diverse and unpredictable terrains underscores the shortcomings of less flexible systems that heavily depend on stable and predictable environments. Moreover, the integration of CBAM~\cite{woo2018cbam} into \textit{TopoNav}'s feature extraction module greatly enhances landmark detection, a common shortcoming in baseline methods. This results in an average 8\% increase in detection accuracy, improving the reliability of subgoal generation and map construction. The subsequent impact on navigation success—up to a 14\% improvement—and reduction in navigation times—up to a 28\% decrease—further establishes \textit{TopoNav} as a significant advancement in autonomous robot navigation.

\begin{table}[ht]
\centering
\scriptsize
\begin{tabular}{|l|c|c|c|c|}
\hline
\textbf{Method} & \textbf{Success} & \textbf{Navigation} & \textbf{Trajectory} & \textbf{Exploration} \\
                & \textbf{Rate (\%)}    & \textbf{Time (s)}       & \textbf{Length (m)}     & \textbf{Coverage (\%)}        \\
\hline
TopoNav & \textbf{98} & \textbf{30.6} & \textbf{10.4} & \textbf{90} \\
\hline
TopoNav & 88 & 40.2 & 14.1 & 80 \\
w/o Topo Map & & & & \\
\hline
TopoNav & 85 & 45.3 & 15.7 & 75 \\
w/o Hierarchy & & & & \\
\hline
TopoNav & 86 & 42.5 & 14.8 & 78 \\
w/o Attention & & & & \\
\hline
\end{tabular}
\caption{Ablation study results.}
\label{tab:ablation}
\end{table}

\subsection{Ablation Study}
To analyze the contribution of each component in \textit{TopoNav}, we conduct an ablation study by removing the topological mapping, the hierarchical structure, and the attention-based CNN separately. Table \ref{tab:ablation} shows the results of the ablation study in the outdoor environment (Scenario 1 \ref{sec:testingscenario}). The results show that removing any component from \textit{TopoNav} decreases success rates and exploration coverage while increasing navigation times and trajectory lengths.


\section{Conclusion, Limitations, and Future Directions}
\label{sec:conclusion}
In this paper, we introduced \textit{TopoNav}, a novel framework for autonomous navigation in unknown environments with sparse rewards. \textit{TopoNav} integrates hierarchical reinforcement learning with dynamic topological mapping to enable efficient exploration, map construction, and goal-directed navigation. The key components of \textit{TopoNav} include a two-level hierarchical policy learning architecture, an active topological mapping approach, and an intrinsic reward mechanism for encouraging exploration. Despite the promising results, \textit{TopoNav} has several limitations that provide an opportunity for future research.  One limitation is the scalability of \textit{TopoNav} to larger and more complex environments. The current implementation may face computational challenges when dealing with extensive topological maps and long-horizon navigation tasks. Future work could explore techniques for map compression and distributed computing to improve the scalability of \textit{TopoNav}. 
Furthermore, developing techniques for distributed map construction, information sharing, and coordinated decision-making could enable more efficient exploration and navigation in large-scale environments with multiple robots.

\section*{Acknowledgment}
This work has been partially supported by the DEVCOM Army Research Laboratory (ARL) under a cooperative agreement (W911NF2120076), ONR Grant \#N00014-23-1-2119, NSF CAREER Award \#1750936, NSF REU Site Grant \#2050999, and NSF CNS EAGER Grant \#2233879.


\bibliographystyle{unsrt}
\bibliography{bibliography}

\end{document}